\documentclass[iicol]{sn-jnl}

\jyear{2021}%

\theoremstyle{thmstyleone}%
\theoremstyle{thmstyletwo}%

\theoremstyle{thmstylethree}%

\begin{document}

\title[Article Title]{Birds' Eye View: Measuring Behavior and Posture of Chickens as a Metric for Their Well-Being.}









\author[1]{\fnm{Kevin Hyekang} \sur{Joo}}\email{hkjoo@umd.edu}

\author[1]{\fnm{Shiyuan} \sur{Duan}}\email{sduan1@umd.edu}

\author[2]{\fnm{Shawna L.} \sur{Weimer}}\email{sweimer@uark.edu}

\author*[1]{\fnm{Mohammad Nayeem} \sur{Teli}}\email{nayeem@umd.edu}

\affil[1]{University of Maryland, College Park, MD}

\affil[2]{University of Arkansas, Fayetteville, AR}









\abstract{Chicken well-being is important for ensuring food security and better nutrition for a growing global human population. In this research, we represent behavior and posture as a metric to measure chicken well-being. With the objective of detecting chicken posture and behavior in a pen, we employ two algorithms: Mask R-CNN for instance segmentation and YOLOv4 in combination with ResNet50 for classification. Our results indicate a weighted F1 score of 88.46\% for posture and behavior detection using Mask R-CNN and an average of 91\% accuracy in behavior detection and 86.5\% average accuracy in posture detection using YOLOv4. These experiments are conducted under uncontrolled scenarios for both posture and behavior measurements. These metrics establish a strong foundation to obtain a decent indication of individual and group behaviors and postures. Such outcomes would help improve the overall well-being of the chickens. The dataset used in this research is collected in-house and will be made public after the publication as it would serve as a very useful resource for future research. To the best of our knowledge no other research work has been conducted in this specific setup used for this work involving multiple behaviors and postures simultaneously.}




\maketitle

\section{Introduction}\label{sec1}

According to the USDA, the total per capita consumption and production of the poultry meat -- specifically chicken meat -- has been growing for decades and is projected to continue to be the dominant meat consumed in the U.S.~\cite{dohlman2020usda}
and the world~\cite{owidmeatproduction}. This growth is in line with the United Nations' goal for sustainable agriculture~\cite{cf2015transforming}. However, for food security and improved nutrition, it is important that, as the production and consumption of chickens increases, the health and the well-being of the animals improve in tandem as there is a direct link between chicken welfare and immunity~\cite{BERGHMAN20162216}.

To meet the growing demands for poultry products and the welfare of the poultry, it is very important to control the spread of diseases,
which requires monitoring chickens and their behavior. However, even broiler companies with years of experience of working with antibiotic-free production systems struggle to control diverse health and disease challenges, ultimately decreasing the welfare of the birds and supply of chicken in the United States~\cite{OVIEDORONDON20191}. Current approaches to evaluate and assess the welfare of chickens traditionally involves human caretaker observation. Based on visual assessments of behavior and body condition, a score is assigned as a health metric~\cite{granquist:19}. However, the welfare of animals also includes affective (mental) state that cannot be assessed by observation of body condition, but can be assessed by observing behavior. Behavior is directly related to the welfare of chickens~\cite{BERGMANN201790}. Human observation will always include inherent inter- and intra-observer variations along with costly and intensive labor for human observation in welfare assessment schemes~\cite{jong:16}. Therefore, there is a need for more innovative robust technological methods to provide accurate information that can assist humans to glean viable on-farm broiler chicken health and welfare outcomes.

Artificial intelligence, specifically computer vision and machine learning, is increasingly playing a bigger role in innovative solutions for animal welfare~\cite{ani6100062,milosevic_ciric_lalic_milanovic_savic_omerovic_doskovic_djordjevic_andjusic_2019,OKINDA2020184}.

In our paper, we employ computer vision techniques to address monitoring chickens for their behavior on a custom new dataset that was also collected by us. Our goal is to observe and analyze chickens' behavior and posture in a pen using a live video stream. 
There are three postures: sitting, standing, and walking. Since two of these postures are stationary, we classify posture as a binary classification problem, with stationary and walking as the two categories.
In addition, we look at the behavior of the chickens, captured through the following behaviors: eating, foraging, preening, allo-preening, drinking, dust-bathing, and control (no behavior or activity). 
Our approach to analyze these postures and behaviors involves: detecting the chickens across frames, identifying one of the postures, and finally associating a behavior with the posture.

\begin{figure*}[ht!]
    \centering
	\includegraphics[width=\linewidth]{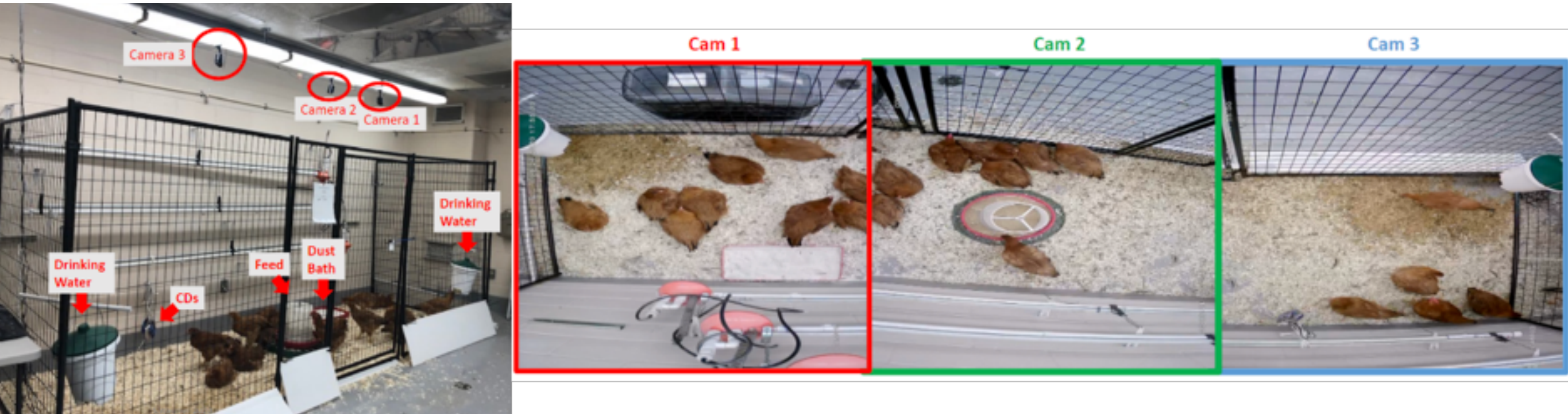} 
\caption{Chickens in the three pens along with the features like the drinking water, feed, dust bath and also the cameras that collect this data.}
\label{pens}       
\end{figure*}

We used a two-pronged approach to localize chickens and detect their posture and behavior. The two approaches include: Mask R-CNN~\cite{kaiming}; and YOLOv4~\cite{YOLOv4} in combination with ResNet-50~\cite{resnet50}. Mask R-CNN detects objects, in our case chickens, and simultaneously generates a high-quality segmentation mask for each instance, which helps determine the posture and behavior as well. However, YOLOv4 is programmed to only localize objects, the chickens in this paper, by providing a bounding box. Our primary rationale for choosing YOLOv4 for the object detection model is that YOLO is well known for its rapidness in computation and effectiveness in detecting non-small (over 32x32 px) objects \cite{small}; the objects we are investigating are mostly considered large (over 96x96 px).  Therefore, in conjunction with the detection model, the ResNet-50 classifier is used to classify the posture and behavior for the particular chicken instance as a follow-up task.
These methods are tested on a new dataset that we collected in a pen of 20 chickens, monitored by three cameras spanning three sections of the same pen. Each of the cameras contained multiple chickens at any given instance. This pen did not have any dividers so birds could freely move between cameras. An image of the pen of chickens is shown in Figure~\ref{pens}. 
The figure also shows the locations of the three cameras that recorded the video. A more detailed description of this dataset is in Section 3.
\section{Related Work}
\subsection{Chicken Disease Monitoring}
Computer vision algorithms have been proven to be suitable for repetitive work, such as identifying diseased chickens~\cite{Bi2018Disease}. In their paper, the authors of previous work have proposed a method of detecting chickens with potential health issues using a non-convolution neural network method. 
Their dataset was composed of 500 images of healthy chickens and 236 images of  diseased chickens, separately captured from the videos taken in the actual feeding environment and classified into either diseased or non-diseased, using a SVM classifier.
Then, combs, eyes and their contours were also obtained in HSV color model by the H threshold. Co-occurrence matrix of H component was calculated from the comb to get the color feature and the texture feature including ASM, COR, IDM, Ent 6 geometry features including A, P, R, E, C and A/P were computed from the eye contour. Finally, all the features mentioned above are fed into an 
SVM classifier to identify diseased chickens.

\subsection{Applications of Mask R-CNN}
Mask R-CNN is a Convolutional Neural Network used for image and instance segmentation. In this work we use it for instance segmentation.  It was built on top of Faster R-CNN~\cite{fasterRCNN} and consists of two stages. The first stage, a Region Proposal Network (RPN), predicts object proposal bounding boxes, and the second stage, an R-CNN detector, refines these proposals, classifies them, and computes pixel-level segmentation for these proposals. This two-stage process makes this approach a combination of object detection, object localization, and object classification, thereby enabling it to clearly distinguish between each object classified as similar instances.

In the domain of behavior recognition for animals and plants, Mask R-CNN has been a popular choice \cite{s19224924,YU2019104846}.  Li \textit{et al.} attempted to observe four pigs in a pen for signs of mounting behavior, which is potentially detrimental to the well-being of pigs. The video of the pigs was recorded in a top-down view, and localized through manual labeling, which was later formatted into JSON files.  Next, the labelled and pre-processed dataset was fed into Mask R-CNN with ResNet-50 as a backbone. 
Beyond behavior recognition, Mask R-CNN has been used for many other problems, including fruit detection \cite{YU2019104846}.  
 In the study of Yu \textit{et al.}, Mask R-CNN, with ResNet-50 as a backbone, was combined with Feature Pyramid Network architecture to localize the fruit picking point and identified ripe fruits from unripe fruits.  
Moreover, Tian \textit{et al.}~\cite{TIAN2020264} created a variation of Mask R-CNN, called MASU R-CNN, which utilizes U-Net as a backbone, to gather detailed information of flowers during their growth.

\subsection{Applications of YOLO}
YOLO, short for You Only Look Once, is a well-known object detector introduced in 2016. It quickly gained popularity for its rapidness and has consistently garnered attention through its continuous development over years \cite{yolo}. One of its family, YOLOv4 is a one-stage object detector built off of the original models~\cite{YOLOv4}. Just like other object detectors, this algorithm has two components -- a backbone and a head. The backbone is pre-trained on ImageNet and encodes relevant information about the input. The head predicts object classes and the bounding box coordinates. YOLOv4 also identifies a ``neck" defined as layers between the backbone and head that serve to collect feature maps from different stages of the network.   
In YOLOv4, the authors combined some of the new features such as, Weighted-Residual-Connections (WRC), Cross-Stage-Partial-connections (CSP), Cross mini-Batch
Normalization (CmBN), Self-adversarial-training (SAT), Mish-activation, Mosaic data augmentation, and CIoU loss, to achieve the state-of-the-art results. 

YOLO has been widely utilized in a variety of fields of study, including poultry science. A paper published by Neethirajan~\cite{chickenYOLO} utilized YOLO as a means of chicken detection. According to the paper, the authors used a dataset comprised of 72 chickens that were from the following three breeds: White Leghorn, Plymouth Rock, and Rhode Island Reds. The chickens were recorded with an RGB camera at varying heights at different times of the day under a variety of lighting conditions. Finally, a YOLO model was trained on the videos collected in a supervised setting and yielded a chicken detection accuracy of over 90\%.  However, the experiment in the paper solely involved chicken detection and did not the behavior recognition of the detected chickens.
In addition, YOLO has also been adopted to detect and classify behaviors of egg-laying breeder chickens~\cite{yolo_egg_breeders}. In the paper, the model was utilized to observe and analyze the behaviors of roosters and hens placed in two wired cages after recording them with cameras placed over the cages. The entire data set consists of 10,230 photos randomly selected in a 30 days period. The sample sets are divided into training (8447 frames), validation (939 frames) and test sets (844 frames). 
Using the data, chickens were classified into one of the six following behaviors with a YOLO model in a supervised manner: mating, standing, feeding, spreading, fighting, and drinking. To our knowledge, this is the closest related work to our research.

\section{Dataset}
In this work, we used a new, custom, in-house-collected video dataset. Videos were collected in May 2020 on twenty male and female broiler chickens (Redbro and Red Ranger, Hubbard, LLC) in a 1.47m wide x 4.45m long pen. Birds were provided a commercial diet \textit{ad libitum} access to feed and water. The pen contained 1 feeder, 2 drinkers, and 2 forms of environmental enrichment: a dust bath (with diatomaceous earth substrate); and CDs hanging from rope (Figure~\ref{pens}). 
The birds had been raised in a separate room in two pens prior to the study. The enrichments were added to the pen on the day prior to video recording. On day 62 of age, 14 continuous hours of video were recorded from 0600-2000 to a DVR from three CCTV video cameras (\textit{HemiVision Model\,\#HM245}). Video cameras were mounted 2.59m above the pen. 

\begin{figure}[ht!]
\centering
	\includegraphics[scale=0.75]{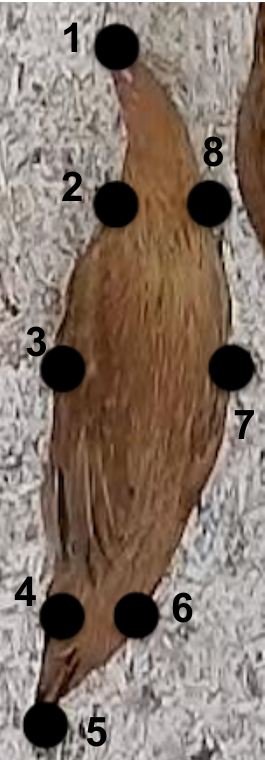} 
\caption{Points used to localize a chicken using LabelMe annotation tool.}
\label{chickenloc}       
\end{figure}

The full dataset contained 504 videos that cumulatively lasted 42 hours. These videos were collected from three different RGB cameras mounted above the same pen in 5-minute clips with a total of 168 videos for each camera. The videos recorded were contiguous. The key distinction between environments is the presence or absence of drinker, feeder, hanging CDs, and dust bath.

In order to create the ground truth data, each one of these videos needed to be labeled frame-by-frame. However, since it is a very laborious task, we annotated 43 videos frame-by-frame.  The annotators had the prior knowledge in poultry science and were trained in detail on poultry behavior by an expert.

The annotations in each frame contain a hand-labeled location as well as posture and behavior information for each chicken. Here, the means of annotations are explained by breaking the process down into two parts. For the first part, each chicken is localized and labeled using LabelMe annotation tool~\cite{russell2008labelme}. This information is then stored in a JSON format. Each chicken is localized by identifying eight points on it as shown in Figure~\ref{chickenloc} to create the ground truth location.

The eight points are annotated in a clockwise direction, and the first and the fifth point represent the head and the tail of the chicken, respectively.  This rule has been carefully honored by the annotators as the location of its head and tail bear a significant meaning in pose estimation, which may be a good new research direction. Our intention is to make the dataset publicly available upon publication for the freedom of different pursuits by other researchers.

Along with the polygonal location information, each chicken is assigned a unique number, a bird ID, at the start of the video that remains consistent across the whole video, for its potential use in tracking in future research. In addition, this identification number is used to associate a posture and a behavior, to be discussed later, with each chicken.

\begin{figure*}[ht!]
    \centering
	\includegraphics[width=0.9\linewidth]{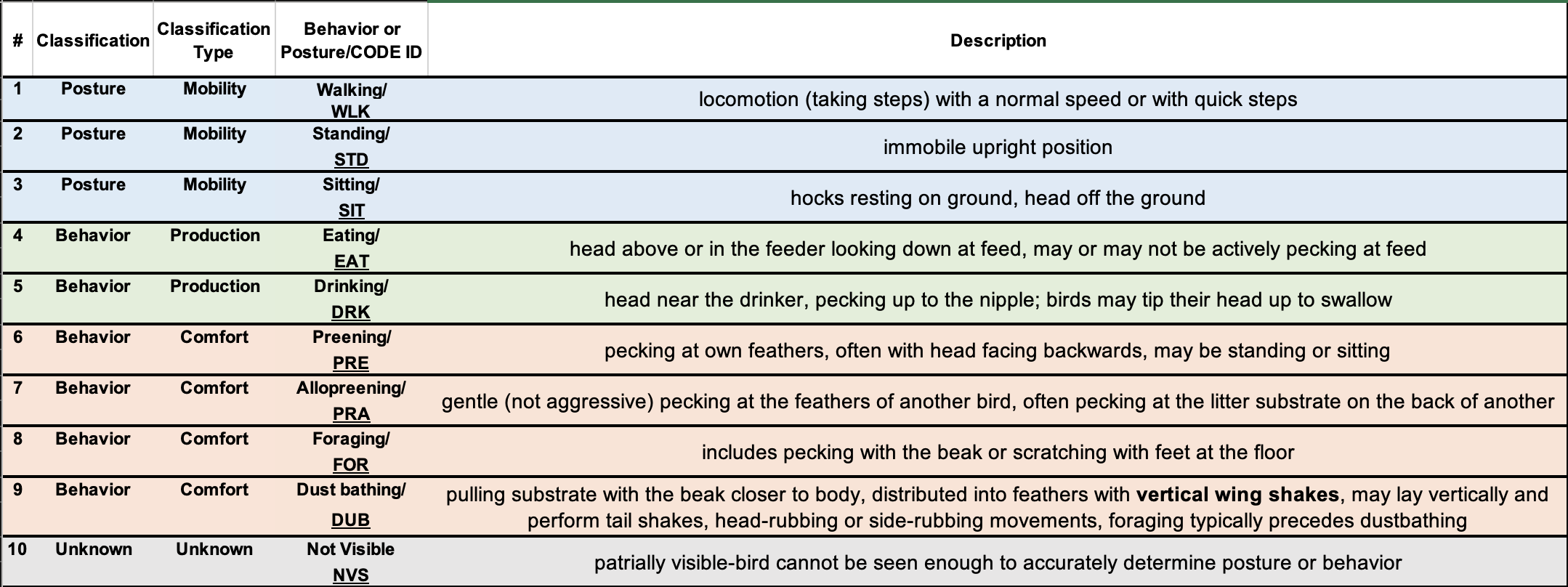} 
\caption{Ethogram: description of the pre-defined postures and behaviors.}
\label{ethogram}       
\end{figure*}

The second part involves an ethogram of pre-defined behaviors as shown in Figure~\ref{ethogram}.
For each localized chicken, the annotator provides the information about its posture and behavior in an excel spreadsheet based on the frame number and the chicken ID.  The spreadsheet is then one-hot encoded into a CSV file. 

\begin{figure*}[ht!]
    \centering
	\includegraphics[width=\linewidth, keepaspectratio]{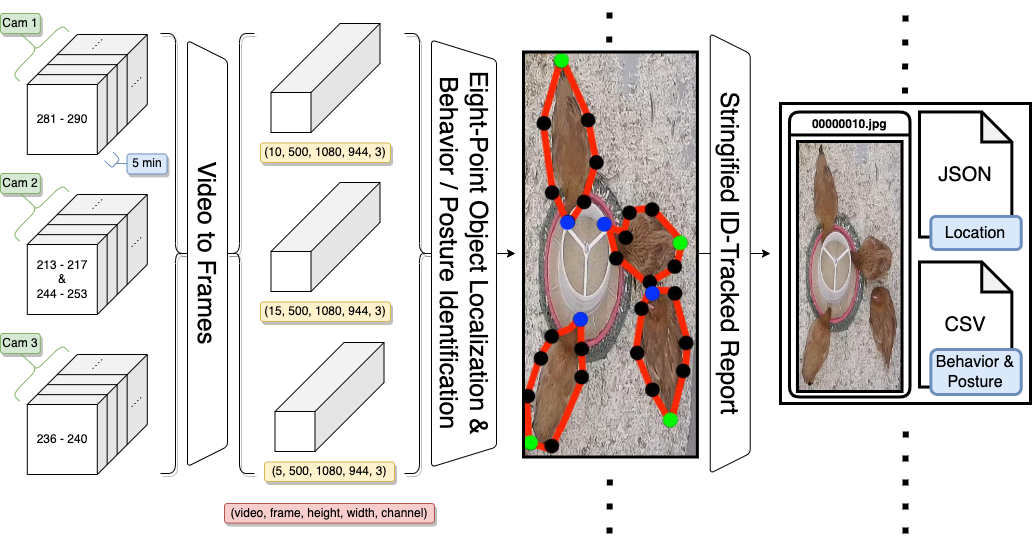} 
\caption{Pre-processing pipeline to obtain the ground truth locations, behavior and posture of the chickens for every frame.}
\label{preprocessing}       
\end{figure*}

\begin{figure*}[ht!]
    \centering
	\includegraphics[width=0.9\linewidth]{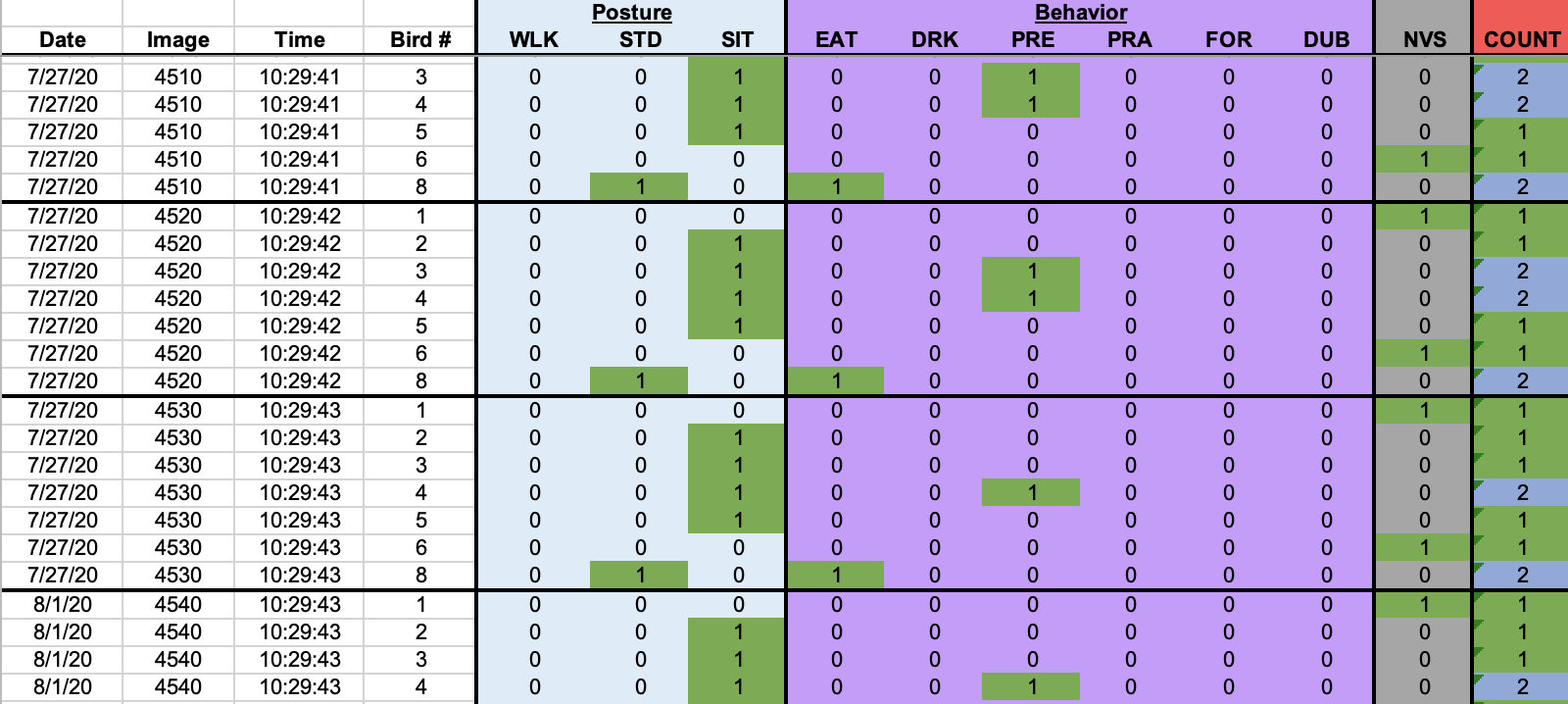} 
\caption{Sample Annotation presenting the various ground truth behaviors and postures of birds in each frame.}
\label{annotation}       
\end{figure*}

Afterward, both the JSON file and CSV file go through a series of sanity checks, such as polygonal coordinate anomaly check. This whole pipeline is demonstrated in Figure~\ref{preprocessing}.
To elaborate on the second part, the posture and behavior spreadsheet is comprised of the following 15 columns (see Figure~\ref{annotation})
: date, image, time, bird ID, walking (WLK), sitting (SIT), standing (STD), eating (EAT), drinking (DRK), preening (PRE), allo-preening (PRA), foraging (FOR), dust-bathing (DUB), not visible (NVS), and count. For all the postures (WLK, SIT, STD) and behaviors (EAT, DRK, PRE, PRA, FOR, DUB) columns, we used 1 to denote that the behavior is associated with the chicken in that particular frame, and 0 otherwise.  
Count represents the sum of posture, behavior, and NVS, and was used for sanity checking (the value must be equal to 1 or 2).

\section{Methods}
In this research, two models were used to detect chickens, and subsequently identify their posture and behavior: Mask R-CNN and YOLOv4 in combination with ResNet50. 
Mask R-CNN algorithm implementation used in this work is part of the Detectron2 framework~\cite{Detectron2018}. 


\subsection{Mask R-CNN}

The chicken dataset is trained in a supervised manner.  The input dataset, in JSON and CSV formats (Figure~\ref{preprocessing}), contains the following information: each chicken's eight-point polygonal coordinates (Figure~\ref{chickenloc}), and labels for posture and behavior. 

\begin{figure}[ht!]
\centering
	\includegraphics[scale=0.6]{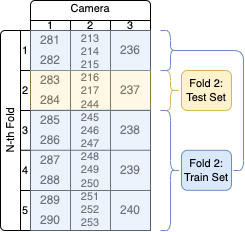} 
\caption{Dividing the camera-split dataset into five folds}
\label{crossvalFolds}       
\end{figure}

Depending on the source, this data is categorized into three groups as follows: videos 281 through 290 from camera 1; videos 213 through 217 \& 244 through 253 from camera 2; videos 236 through 240 from camera 3. Then, each dataset, which has been split by camera number, is divided into five folds for cross-validation (Figure~\ref{crossvalFolds}).

For example, the camera 1's four folds 1, 3, 4, and 5 were fed into Detectron2 for training, and the remaining fold 2 was used for testing.  Afterwards, the result was averaged across folds by camera number.  

\begin{figure}[ht!]
\centering
	\includegraphics[width=\linewidth]{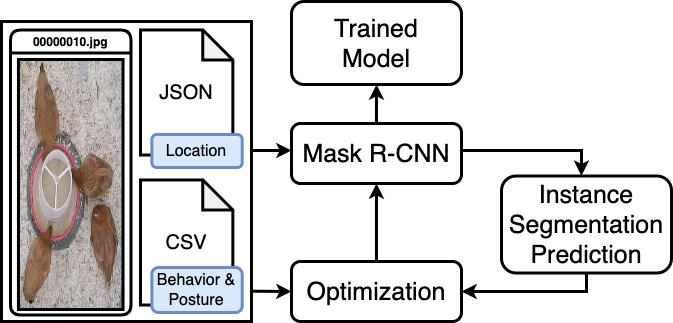} 
\caption{Mask R-CNN pipeline}
\label{maskrcnn}       
\end{figure}

In summary, a total of fifteen experiments (i.e., three cameras * five folds) were conducted, and three averages reported, one for each camera.  We trained two separate cross-validation models, one for posture and the other for behavior classification. Figure~\ref{maskrcnn} displays a Mask R-CNN pipeline demonstrating the application of the algorithm for instance segmentation on our chicken dataset. 

\begin{figure}[ht!]
\centering
	\includegraphics[width=\linewidth]{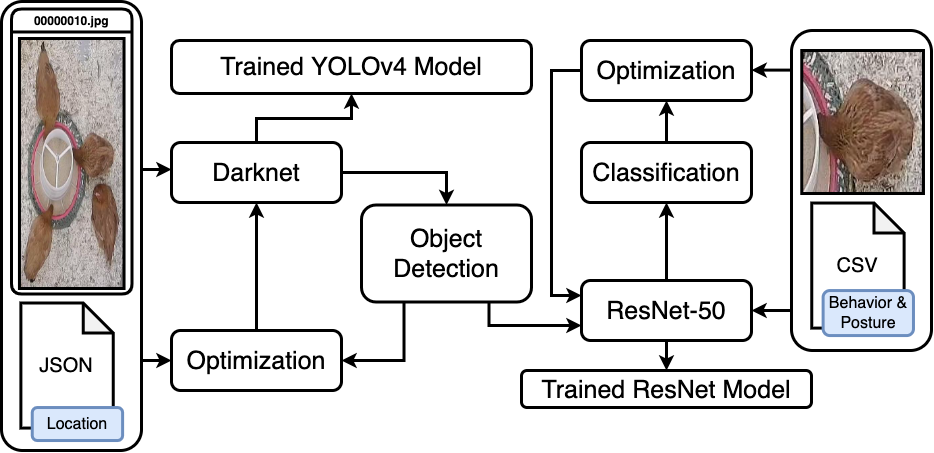} 
\caption{YOLO \& ResNet pipeline}
\label{yoloresnet}       
\end{figure}



\subsection{YOLOv4}

We use YOLOv4 object detector to localize chickens much like our experiments using Mask R-CNN, albeit without labels for behavior or posture. Similar to the previous experiments, the chicken dataset consists of the actual frames that contain the chickens along with the JSON files describing the bounding box of each chicken. The chickens are localized using the 8 polygonal coordinates previously described. 
We performed two 5-fold cross-validation experiments. In all, we ran 10 experiments, each with a different training and testing set of videos as presented in Table~\ref{yolocrossval}. 

\begin{table}[]
\centering
\caption{5-fold cross-validated average precision of YOLOv4 in different IoU thresholds. }
\begin{tabular}{ccccc}
\hline
\multicolumn{1}{l}{}      & Test Videos & AP:0.1 & AP:0.5 & AP:0.75 \\ \hline
\multirow{5}{*}{Camera 2} & 213 214 215 & 0.93     & 0.85     & 0.48      \\
                          & 216 217 244 & 0.93     & 0.84     & 0.53      \\
                          & 245 246 247 & 0.90     & 0.78     & 0.64      \\
                          & 248 249 250 & 0.94     & 0.83     & 0.57      \\
                          & 251 252 253 & 0.93     & 0.82     & 0.55      \\ \hline
\multirow{5}{*}{Camera 1} & 281 282     & 0.91     & 0.87     & 0.47      \\
                          & 283 284     & 0.92     & 0.88     & 0.48      \\
                          & 285 286     & 0.91     & 0.88     & 0.46      \\
                          & 287 288     & 0.90      & 0.87      & 0.46       \\
                          & 289 290     & 0.91      & 0.87      & 0.48       \\ \hline
\end{tabular}
\label{yolocrossval}
\end{table}

However, unlike Mask R-CNN, YOLOv4 only returns the chicken locations. Since our goal is to detect chicken behaviors and postures, YOLOv4 is not suitable by itself. 
For this research, we use ResNet-50 for both behavior and posture classifications, post-object detection using YOLO. The complete pipeline of our solution using YOLOv4 in combination with the ResNet-50 classifier is presented in Figure~\ref{yoloresnet} and described in the following section.


\subsubsection{Chicken classifiers}
The output of the object detection is the localization of the chicken in the image frames. Each of these detected chickens is cropped and then processed for posture and behavior classifications, separately. 

The behavior classifier takes the YOLO-detected chicken and classifies it as one of the five classes: control, eating, preening, allo-preening, and foraging. The posture detector is a binary classifier that classifies each chicken into either stationary or walking. Stationary chickens are either sitting or standing. 

We evaluated the bounding boxes detected by YOLO against the ground truth bounding boxes based on the Intersection over Union (IoU) score between the two. The ground truth bounding boxes are stored in an input CSV file along with their labels. With this dataset, we are able to train our classifiers in a supervised-learning manner. 

Our classifiers have 4 major components: the preprocessing layer, the feature extractor, the location embedding layer, and the prediction layer. The preprocessing layer normalizes the input image and also augments the data using random rotations (random horizontal and vertical flips are also included). The backbone is a ResNet-50 feature extractor that is pretrained with ImageNet dataset. The location embedding layer is a 3 layer MLP (Multi-Layer-Perceptron) that takes four 2-dimensional vectors representing the bounding box location with respect to the entire frame and encodes these vectors into a 16-dimensional vector. This 16-dimensional vector is concatenated with the output vector of the feature extractor and fed into the prediction layer. 
Both posture and behavior classifiers are cross-validated in different train-test-split settings. Each experiment has different train and test videos. The average of train accuracy, test accuracy, and unweighted F1 scores (macro F1) are calculated and evaluated. 
Our dataset has an imbalance in the number of different behaviors. To investigate this we also performed class-wise evaluation experiments.

\section{Results}
In this section we present the experimental results for the two algorithms discussed in the previous sections. Firstly, we present the posture and behavior results of Mask R-CNN experiments. Secondly,  we present the detection accuracy of the YOLOv4 model we trained using Darknet library~\cite{darknet13}, followed by the results of behavior and posture classification. In addition, we also present the model test accuracy after cross validation and F1 scores (weighted and unweighted). For each classifier, a precision and recall curve is also presented. 

\subsection{Evaluation Metric}
Object detection accuracy is calculated using the IoU metric.  Each ground-truth object is paired with one of the prediction bounding boxes that has the highest IoU out of the ones whose IoU is over a certain threshold; the same prediction bounding box is not used to pair multiple times. The process can be summarized in the equation below:

\begin{equation}
\begin{gathered}
    \max_{j} \text{ IoU}(gt_i, pred_j) \text{\;\;\;s.t.\;} \text{ IoU} > \alpha 
\end{gathered}
\end{equation}

where $\alpha$ is an IoU threshold ranging between 0 and 1, \textit{gt} is the ground truth chicken, \textit{pred} is one of \textit{j} chickens found to have an IoU of at least $\alpha$; duplicates ignored, whichever has the highest IoU among \textit{j} chickens will be paired with \textit{i}-th \textit{gt} chicken, while rejecting the duplicate IoU pairing.

\subsection{Mask R-CNN}
The performance of the model was evaluated using the popular method: COCOEvaluation. COCOEvaluation works such that it first measures the AP at 10 different intervals -- from IoU threshold of 50\% until 95\% at an increment of 5\% -- and takes the average of its performance at varying levels.  However, instead of reporting solely the average, we decided to report its AP result for the entire 10 intervals for careful comparison. 

We picked two models that effectively reflect the performance of our general models while also capturing as much categorical information as possible. The representative selections include, Camera 1 video with Sequence 283-284 as the test set (\textit{i.e.,} Fold 2 of Camera 1 from Figure~\ref{crossvalFolds}) and a Camera 2 video with Sequence 248-250 as the test set (\textit{i.e.,} Fold 4 of Camera 2 from Figure~\ref{crossvalFolds}).    

\begin{figure}[ht!]
\centering
	\includegraphics[width=0.9\linewidth]{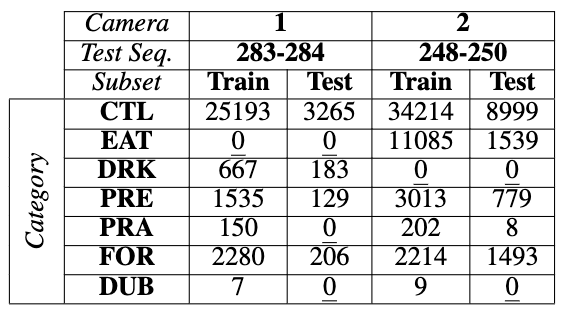} 
\caption{Categorical distribution of the datasets used in Figure \ref{test283284} and Figure \ref{test248249250}}
\label{metadata}       
\end{figure}

The metadata provided in Fig. \ref{metadata} helps convey the categorical distribution of each dataset. The numbers in Fig. \ref{metadata} represent the ground-truth counts of chickens for each behavior \textit{category} in the corresponding \textit{subset} of the dataset. It is important to note that, if there is no instance of a chicken exhibiting a particular behavior in the test video sequence (\textit{i.e,} a cell in the test column contains 0), the line for the behavior is not plotted in the graph.

\begin{figure}[ht!]
\centering
	\includegraphics[width=0.9\linewidth]{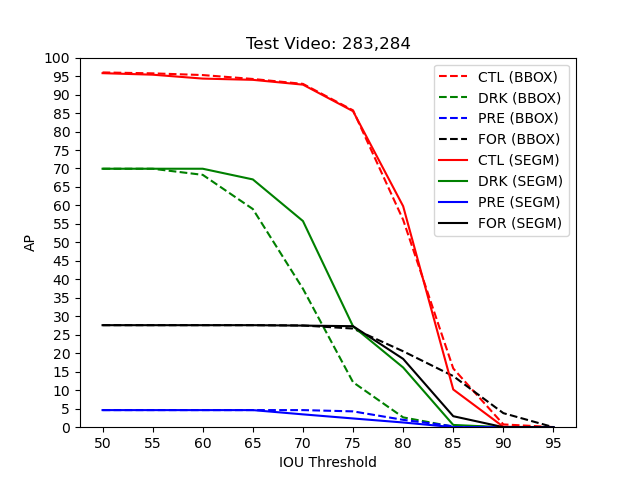} 
\caption{Average Precision with respect to an IoU Threshold}
\label{test283284}       
\end{figure}

\begin{figure}[ht!]
\centering
	\includegraphics[width=0.9\linewidth]{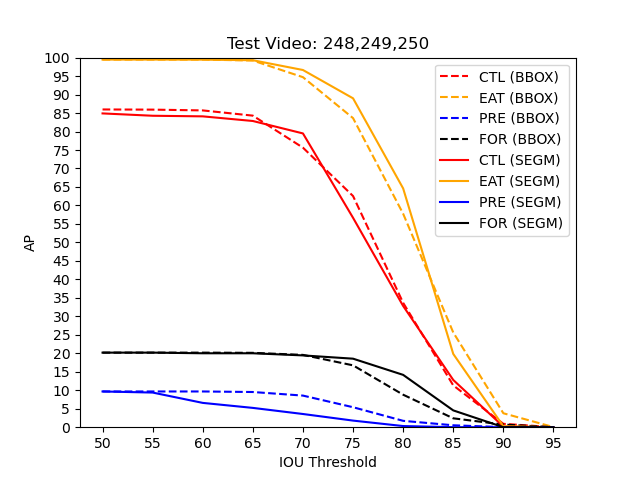} 
\caption{Average Precision with respect to an IoU Threshold}
\label{test248249250}       
\end{figure}

For every category, We present the performance outcomes using a rectangular boundary-box (bbox) and segmentation results of an object detector in Figures~\ref{test283284} and \ref{test248249250}. The segmentation information is plotted in solid lines, and bbox information in dotted lines.

It can be observed from both graphs that segmentation approach does not necessarily outperform the bbox method encompassing the background, although they both perform competitively well.  In addition, we can mostly see the plateau at the IoU threshold of 75\%.

\subsection{YOLOv4}
The performance of YOLOv4 is evaluated based on the Average Precision (AP) given different IoU thresholds. Results of a five-fold cross-validated experiment is shown in Table~\ref{yolocrossval}. We trained 5 models for both camera 1 and camera 2. Each model is trained on different sets of train and test videos. Our model is proven to have an average precision of 85\% when IoU is set to 0.5. This indicates that when the overlap between the bounding boxes of the YOLO detected chickens and the actual bounding box of chickens is at least 50\%, the average precision for detection is between 78-85\% for Camera 2 and around 87\% for the Camera 1. 

\subsubsection{Behavior and Posture classification}
\begin{figure}[h!]
	\includegraphics[width=\linewidth]{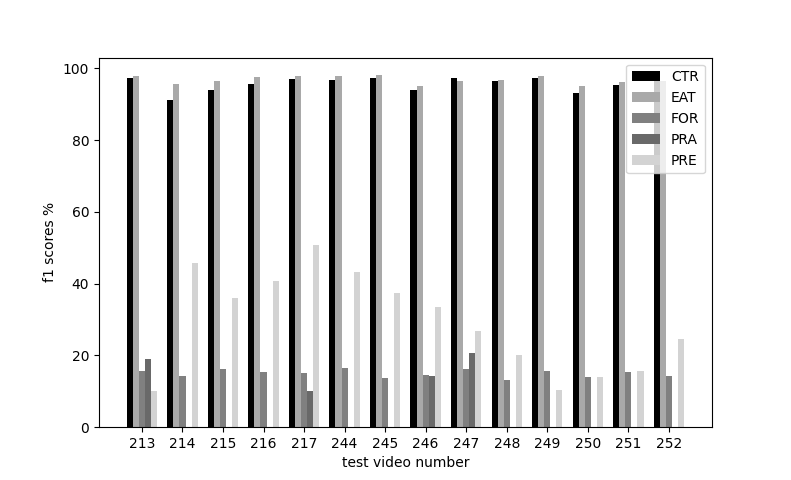} 
\caption{Class-wise F1 scores of behavior classifier}
\label{CWF1}       
\end{figure}

\begin{figure}[ht!]
	\includegraphics[scale=0.5]{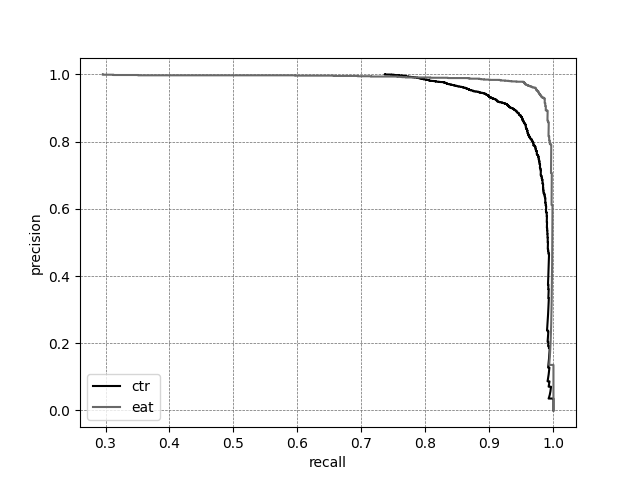} 
\caption{Behavior classifier PR curve for eating and control}
\label{BPR}       
\end{figure}

Table~\ref{yoloposres} shows the test accuracy and F1 scores for posture classifier. 
\begin{table}[]
\caption{This table shows accuracy and F1 scores for the posture detector on test videos}
\begin{tabular}{cccc}
\hline
Test Video \#     & Accuracy  & Weighted F1   & Unweighted F1\\ \hline
213 214 215      & 88.09\%   & 88.46\%       & 82.70\% \\ 
216 217 244      & 85.32\%   & 87.45\%       & 81.95\% \\ 
213 214 215      & 86.55\%   & 88.11\%       & 82.05\% \\ \hline
\end{tabular}
\label{yoloposres}
\end{table}
The best test accuracy reached by our model is 88.09\% with a weighted F1 score of 88.46\%. Table~\ref{yolobehres} displays the test accuracy and macro F1 scores of the behavior detector. 
\begin{table}[]
\caption{This table shows the cross-validation results for the behavior detector}
\begin{tabular}{cccc}
\hline
{ Test Video} & Train Accuracy & Test Accuracy & Macro F1 \\ \hline
213                                                       & 99.34\%        & 95.31\%       & 47.96\%  \\ 
214                                                       & 99.50\%        & 89.61\%       & 52.93\%  \\ 
215                                                       & 99.15\%        & 87.94\%       & 49.37\%  \\ 
216                                                       & 99.31\%        & 91.19\%       & 49.43\%  \\ 
217                                                       & 99.24\%        & 92.17\%       & 51.29\%  \\ 
244                                                       & 99.27\%        & 93.12\%       & 47.29\%  \\ 
245                                                       & 99.52\%        & 94.41\%       & 50.91\%  \\ 
246                                       & 99.35\%        & 90.12\%       & 48.83\%  \\ 
247                                       & 99.41\%        & 92.33\%       & 50.01\%  \\ 
248                                       & 99.50\%        & 89.81\%       & 49.80\%  \\ 
249                                       & 99.36\%        & 90.74\%       & 50.19\%  \\ 
250                                       & 99.47\%        & 91.29\%       & 49.97\%  \\ 
251                                       & 99.27\%        & 89.19\%       & 49.76\%  \\ 
252                                                       & 99.27\%        & 89.19\%       & 49.76\%  \\ \hline
Average                                                   & 99.36\%        & 91.18\%       & 49.83\%  \\ \hline
\end{tabular}
\label{yolobehres}
\end{table}
The best test accuracy is 95.31\% and the best macro F1-score is 52.93\%. The macro F1 score is considerably lower than the accuracy because our dataset is highly unbalanced, and this problem can be better visualized in Figure \ref{CWF1}. Table \ref{classwisef1} and 
Figure~\ref{CWF1} shows the class-wise F1-scores for each test videos. The average F1-scores are 95.76\%, 96.85\%, 15.01\%, 4.57\% 29.22\% for CTR, EAT, FOR, PRA, PRE, respectively.
The discrepancy is explained due to the imbalance in the various categories of the behaviors. For an experiment like this the F1 score results are much more reliable than the accuracy scores. The low F1 score is owing to the fact that the number of examples of FOR, PRE and PRA are far lower than that of the CTR and EAT categories. Due to the low precision and recall of the first three categories, the overall F1 score is lowered. However, the effectiveness of the overall approach is demonstrated by the precision-recall curves for two largest behavior categories, EAT and CTR, in Figure~\ref{BPR}.

\begin{table}[]
\centering
\caption{This table shows class-wise F1 scores of behavior detector across 5 different classes}
\label{classwisef1}
\scalebox{0.89}{
\begin{tabular}{cccccc}
\hline
\textbf{Test Video} & \textbf{CTR}     & \textbf{EAT}     &\textbf{FOR}     & \textbf{PRA}     & \textbf{PRE}     \\ \hline
213                                                       & 97.24\% & 97.85\% & 15.68\% & 19.04\% & 10.00\% \\ 
214                                  & 91.28\% & 95.67\% & 14.28\% & 0.00\%  & 45.90\% \\ 
215                                  & 94.05\% & 96.49\% & 16.21\% & 0.00\%  & 36.00\% \\ 
216                                  & 95.66\% & 97.72\% & 15.25\% & 0.00\%  & 40.70\% \\ 
217                                  & 97.11\% & 97.84\% & 15.19\% & 10.00\% & 50.72\% \\ 
244                                  & 96.72\% & 97.88\% & 16.44\% & 0.00\%  & 43.27\% \\ 
245                                  & 97.47\% & 98.12\% & 13.82\% & 0.00\%  & 37.27\% \\ 
246                                  & 93.95\% & 95.22\% & 14.63\% & 14.28\% & 33.63\% \\ 
247                                  & 97.26\% & 96.45\% & 16.31\% & 20.72\% & 26.73\% \\ 
248                                  & 96.52\% & 96.77\% & 13.25\% & 0.00\%  & 20.14\% \\ 
249                                  & 97.28\% & 97.94\% & 15.62\% & 0.00\%  & 10.25\% \\ 
250                                  & 93.26\% & 95.22\% & 13.91\% & 0.00\%  & 13.95\% \\ 
251                                  & 95.51\% & 96.24\% & 15.26\% & 0.00\%  & 15.78\% \\ 
252                                                       & 97.27\% & 96.44\% & 14.29\% & 0.00\%  & 24.70\% \\ \hline
Average                                                   & 95.76\% & 96.85\% & 15.01\% & 4.57\%  & 29.22\% \\ \hline
\end{tabular}}
\end{table}

\section{Discussion}
In this research we present behavior and posture detection as a metric to measure the  behavioral well-being of chickens. In order to obtain behaviors and postures, we employ two approaches: Mask R-CNN and YOLOv4 with ResNet50. Mask R-CNN is an instance segmentation object detector to obtain the behavior and posture of chickens. YOLOv4 is an object detector that is used in combination with ResNet50 to obtain the behaviors and postures. 

The contributions of this research include:
\begin{itemize}
    \item collection of a new custom dataset to measure the behavior of chickens in a pen using multiple cameras.
    \item behavior and posture detection with high accuracy by applying the state-of-the art algorithms.
    \item laying a foundation for measuring chicken well-being through posture and behavior metrics.
    \item laying a foundation for understanding group behavior of the chickens.
\end{itemize}

Our results indicate a high performance of these two approaches for behavior and posture detection of chickens despite the challenges involved. There are many challenges in a problem like this, and we will discuss some of those here. Firstly, to our knowledge, this is the only object detection problem in which the object is changing its size with each day; secondly, the objects bear uncanny resemblance; thirdly, the behaviors can be misinterpreted in a setting like this where multiple chickens are in close proximity. For example, preening and allo-preening as well as standing and sitting can easily be misinterpreted. That is the reason we had to combine sitting and standing as a single posture. Finally, since the chicken behavior cannot be controlled, there is a huge imbalance between different types of postures and behaviors. 
This problem and its impact is clearly indicated in Figure~\ref{CWF1}. The macro-F1 scores are greatly influenced by the number of examples available in the training data. Since there are fairly reasonable number of examples of eating and no behavior, our approaches are reasonably accurate.  
Measuring behaviors and postures of individual chickens would help mark the behavior and the posture of a group of chickens. For example, when we notice most of the chickens in a pen are not drinking enough water or not eating much, this would indicate a bigger problem that would require intervention and hence lead to better well being of chickens.
\bmhead{Acknowledgments}
This work was conducted when Shawna Weimer was at the University of Maryland, College Park. The data collection and labeling of the project was supported by the startup funds provided to Shawna by
the University of Maryland Department of Animal and Avian
Sciences. We would like to acknowledge Seungwon Oh and Ali Atai for their initial work in the project and Ashanti Mangrum, Heidi Rinehart, Maryann Khong, and Noelle Potter for their work in labeling the data set used in this research.

\bibliographystyle{plain}
\bibliography{refs}

\begin{thebibliography}{10}

\bibitem{ani6100062}
Neila Ben~Sassi, Xavier Averós, and Inma Estevez.
\newblock Technology and poultry welfare.
\newblock {\em Animals}, 6(10), 2016.

\bibitem{BERGHMAN20162216}
L.R. Berghman.
\newblock Immune responses to improving welfare.
\newblock {\em Poultry Science}, 95(9):2216--2218, 2016.

\bibitem{BERGMANN201790}
Shana Bergmann, Angela Schwarzer, Katharina Wilutzky, Helen Louton, Josef
  Bachmeier, Paul Schmidt, Michael Erhard, and Elke Rauch.
\newblock Behavior as welfare indicator for the rearing of broilers in an
  enriched husbandry environment—a field study.
\newblock {\em Journal of Veterinary Behavior}, 19:90--101, 2017.

\bibitem{Bi2018Disease}
M.~Bi, T.~Zhang, X.~Zhuang, and P.~Jiao.
\newblock Recognition method of sick yellow feather chicken based on head
  features.
\newblock {\em Nongye Jixie Xuebao/Transactions of the Chinese Society for
  Agricultural Machinery}, 49:51--57, 01 2018.

\bibitem{YOLOv4}
Alexey Bochkovskiy, Chien-Yao Wang, and Hong-Yuan~Mark Liao.
\newblock Yolov4: Optimal speed and accuracy of object detection, 2020.

\bibitem{cf2015transforming}
ODDS Cf.
\newblock Transforming our world: the 2030 agenda for sustainable development.
\newblock {\em United Nations: New York, NY, USA}, 2015.

\bibitem{jong:16}
I.~C. de~Jong, V.~A. Hindle, A.~Butterworth, B.~Engel, P.~Ferrari, H.~Gunnink,
  T.~Perez~Moya, F.~A. Tuyttens, and C.~G. van Reenen.
\newblock Simplifying the welfare quality assessment protocol for broiler
  chicken welfare.
\newblock {\em Animal : an international journal of animal bioscience},
  10(1):117--127, 2016.

\bibitem{dohlman2020usda}
Erik Dohlman, James Hansen, David Boussios, et~al.
\newblock Usda agricultural projections to 2029.
\newblock {\em USDA agricultural projections to 2029.}, 2020.

\bibitem{Detectron2018}
Ross Girshick, Ilija Radosavovic, Georgia Gkioxari, Piotr Doll\'{a}r, and
  Kaiming He.
\newblock Detectron.
\newblock \url{https://github.com/facebookresearch/detectron}, 2018.

\bibitem{granquist:19}
E.~G. Granquist, G.~Vasdal, I.~C. de~Jong, and R.~O. Moe.
\newblock Lameness and its relationship with health and production measures in
  broiler chickens.
\newblock {\em Animal : an international journal of animal bioscience},
  13(10):2365--2372, 2019.

\bibitem{kaiming}
Kaiming He, Georgia Gkioxari, Piotr Dollár, and Ross Girshick.
\newblock Mask r-cnn.
\newblock In {\em 2017 IEEE International Conference on Computer Vision
  (ICCV)}, pages 2980--2988, 2017.

\bibitem{resnet50}
Kaiming He, Xiangyu Zhang, Shaoqing Ren, and Jian Sun.
\newblock Deep residual learning for image recognition, 2015.

\bibitem{small}
Mate Kisantal, Zbigniew Wojna, Jakub Murawski, Jacek Naruniec, and Kyunghyun
  Cho.
\newblock Augmentation for small object detection, 2019.

\bibitem{s19224924}
Dan Li, Yifei Chen, Kaifeng Zhang, and Zhenbo Li.
\newblock Mounting behaviour recognition for pigs based on deep learning.
\newblock {\em Sensors}, 19(22), 2019.

\bibitem{milosevic_ciric_lalic_milanovic_savic_omerovic_doskovic_djordjevic_andjusic_2019}
B.~Milosevic, S.~Ciric, N.~Lalic, V.~Milanovic, Z.~Savic, I.~Omerovic,
  V.~Doskovic, S.~Djordjevic, and L.~Andjusic.
\newblock Machine learning application in growth and health prediction of
  broiler chickens.
\newblock {\em World's Poultry Science Journal}, 75(3):401–410, 2019.

\bibitem{chickenYOLO}
Suresh Neethirajan.
\newblock Chicktrack - a quantitative tracking tool for measuring chicken
  activity.
\newblock {\em Measurement}, page 110819, 2022.

\bibitem{OKINDA2020184}
Cedric Okinda, Innocent Nyalala, Tchalla Korohou, Celestine Okinda, Jintao
  Wang, Tracy Achieng, Patrick Wamalwa, Tai Mang, and Mingxia Shen.
\newblock A review on computer vision systems in monitoring of poultry: A
  welfare perspective.
\newblock {\em Artificial Intelligence in Agriculture}, 4:184--208, 2020.

\bibitem{OVIEDORONDON20191}
Edgar~O. Oviedo-Rondón.
\newblock Holistic view of intestinal health in poultry.
\newblock {\em Animal Feed Science and Technology}, 250:1--8, 2019.
\newblock SI: Intestinal Health.

\bibitem{darknet13}
Joseph Redmon.
\newblock Darknet: Open source neural networks in c.
\newblock \url{http://pjreddie.com/darknet/}, 2013--2016.

\bibitem{yolo}
Joseph Redmon, Santosh Divvala, Ross Girshick, and Ali Farhadi.
\newblock You only look once: Unified, real-time object detection, 2015.

\bibitem{fasterRCNN}
Shaoqing Ren, Kaiming He, Ross Girshick, and Jian Sun.
\newblock Faster r-cnn: Towards real-time object detection with region proposal
  networks.
\newblock {\em IEEE Transactions on Pattern Analysis and Machine Intelligence},
  39(6):1137--1149, 2017.

\bibitem{owidmeatproduction}
Hannah Ritchie and Max Roser.
\newblock Meat and dairy production.
\newblock {\em Our World in Data}, 2017.
\newblock https://ourworldindata.org/meat-production.

\bibitem{russell2008labelme}
Bryan~C Russell, Antonio Torralba, Kevin~P Murphy, and William~T Freeman.
\newblock Labelme: a database and web-based tool for image annotation.
\newblock {\em International journal of computer vision}, 77(1-3):157--173,
  2008.

\bibitem{TIAN2020264}
Yunong Tian, Guodong Yang, Zhe Wang, En~Li, and Zize Liang.
\newblock Instance segmentation of apple flowers using the improved mask
  r–cnn model.
\newblock {\em Biosystems Engineering}, 193:264--278, 2020.

\bibitem{yolo_egg_breeders}
Juan Wang, Nan Wang, Lihua Li, and Zhenhui Ren.
\newblock Real-time behavior detection and judgment of egg breeders based on
  yolo v3.
\newblock {\em Neural Computing and Applications}, 2019.

\bibitem{YU2019104846}
Yang Yu, Kailiang Zhang, Li~Yang, and Dongxing Zhang.
\newblock Fruit detection for strawberry harvesting robot in non-structural
  environment based on mask-rcnn.
\newblock {\em Computers and Electronics in Agriculture}, 163:104846, 2019.

\end{thebibliography}


\end{document}